\documentclass{article}

\def\arxivmode{}
\ifdefined\arxivmode
  \usepackage[preprint]{neurips_2026}
\else
  \usepackage{neurips_2026}
\fi
\newcommand{\projecturl}{http://circle-group.github.io/research/SaMoR}
\usepackage[utf8]{inputenc}
\usepackage[T1]{fontenc}
\usepackage{microtype}
\usepackage{url}
\usepackage{hyperref}
\usepackage{enumitem}

\usepackage{amsmath}
\usepackage{amsfonts}
\usepackage{amssymb}
\usepackage{bm}
\usepackage{nicefrac}
\usepackage{pifont}

\usepackage{booktabs}
\usepackage{multirow}
\usepackage{graphicx}
\usepackage{subcaption}
\usepackage[dvipsnames]{xcolor}

\usepackage{cleveref}
\usepackage{xspace}

\renewcommand{\paragraph}[1]{{\vspace{0.3mm}\noindent \bf #1}}

\crefname{figure}{Fig.}{Figs.}
\Crefname{figure}{Fig.}{Figs.}
\crefname{table}{Table}{Tables}
\Crefname{table}{Table}{Tables}
\crefname{section}{Sec.}{Secs.}
\Crefname{section}{Sec.}{Secs.}
\crefname{equation}{Eq.}{Eqs.}
\Crefname{equation}{Eq.}{Eqs.}

\title{SAMoR: Motion Modelling for Articulated Objects\\of Any Skeleton and Topology}

\author{%
  Yuhao Zhang\\
  {\small Imperial College London}
  \And
  Gerard Pons-Moll\\
  {\small University of T\"ubingen, T\"ubingen AI Center}
  \And
  Tolga Birdal\\
  {\small Imperial College London}
}

\begin{document}

\maketitle

% ---------------------------------------------------------------------
% Abstract
% ---------------------------------------------------------------------
\begin{abstract}
  Modeling motion for articulated objects of arbitrary skeleton topology remains difficult: existing motion generators target a fixed human skeleton, and prior adaptations either fail to share a vocabulary across rigs or discard motion detail through global pooling. Our key observation is that while joint-level motion does not correspond cleanly across species, motion of functional joint groups does---a human arm, a wolf foreleg, and a bird wing share semantic motion structure despite differences in joint count and connectivity, a correspondence that joint names (e.g., ``forearm'', ``wing\_L1'') partially expose even when topology does not. We introduce SAMoR (Skeleton-Aware Motion Representation for Articulated Objects), a cross-topology motion representation that encodes each motion segment as a small fixed number ($K{=}8$) of part tokens shared across arbitrary skeletons. SAMoR takes three input signals---per-joint motion features, kinematic graph structure, and joint-name embeddings---and processes them with a graph-transformer encoder, then compresses the resulting heterogeneous per-joint features into part-level tokens via cross-attention pooling and residual vector quantization, yielding a discrete motion codebook shared across rigs. To prevent the part queries from collapsing into redundant global representations, we introduce a topology-agnostic attention supervision loss, combined with random joint-name dropout to prevent over-reliance on text labels; together these encourage the part tokens to cluster joints into functional groups from names, structure, and motion jointly. We curate a unified heterogeneous motion corpus from HumanML3D, Truebones Zoo, and animated Objaverse-XL assets, and evaluate SAMoR on held-out characters with unseen skeletons. The resulting representation supports accurate reconstruction and cross-topology motion transfer, and further enables text-conditioned generation and localized part-wise editing through a MaskGIT token generator. SAMoR reaches $2.75\!\times\!10^{-2}$ normalized MPJPE on cross-topology reconstruction---$5.8\!\times$ below the strongest adapted variable-$J$ tokenizer baseline---and remains competitive with fixed-skeleton specialists on HumanML3D for both VQ-VAE reconstruction and text-to-motion generation. Our code and extended results can be found at \url{\projecturl}.

\end{abstract}

% ---------------------------------------------------------------------
% Main body  (shared with main.tex)
% ---------------------------------------------------------------------
%\input{sec/1_intro}
\section{Introduction}
\label{sec:intro}

\begin{figure}[!t]
  \centering
  \includegraphics[width=\linewidth]{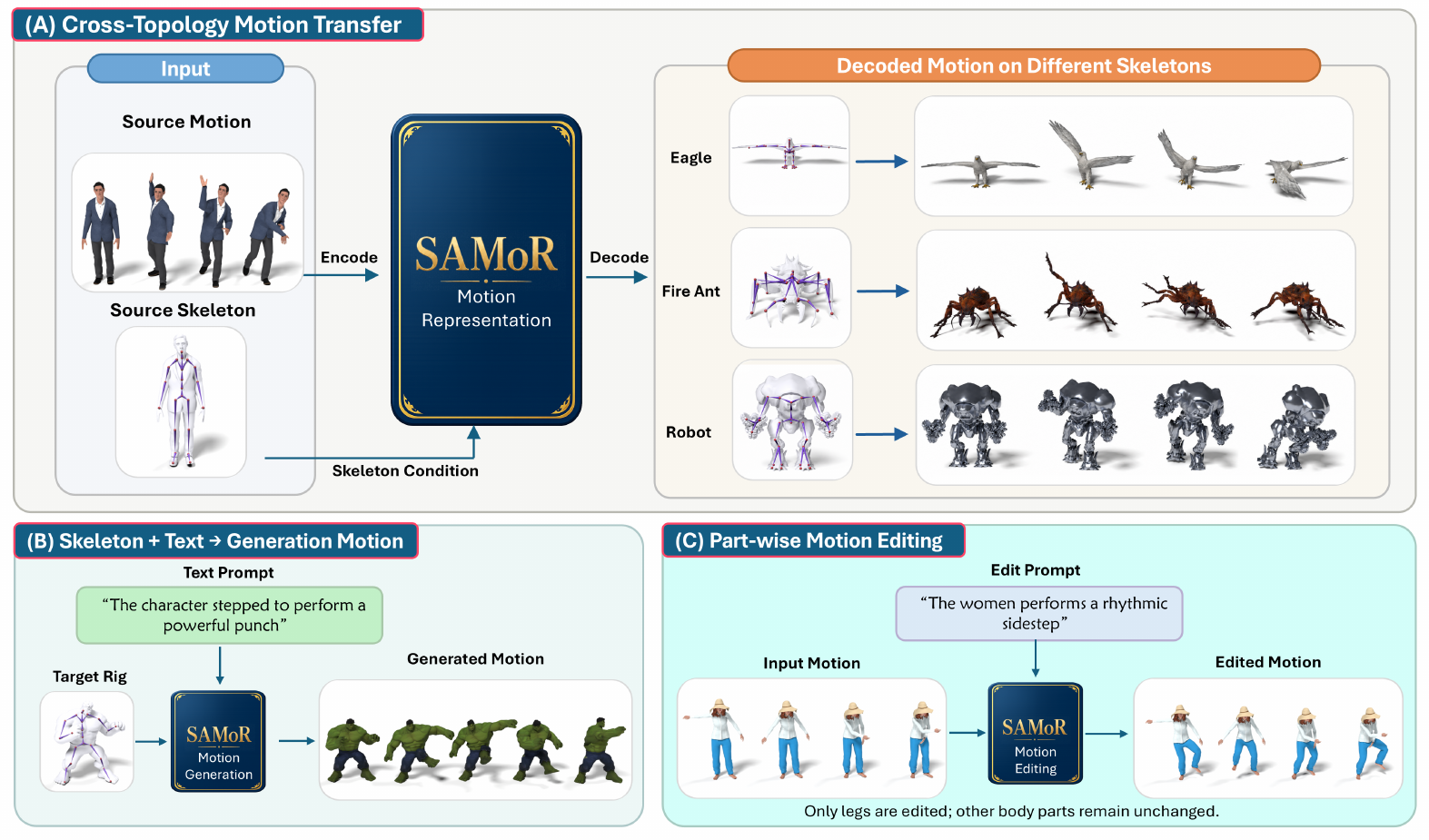}
  \caption{\textbf{SAMoR enables three downstream tasks
    from a single cross-topology motion representation.}
    \textbf{(A)} Cross-topology motion transfer by swapping the
    skeleton conditioning at decode time (eagle, fire ant, robot).
    \textbf{(B)} Text-to-motion generation conditioned on a target rig.
    \textbf{(C)} Part-wise editing: replacing selected part tokens edits
    only the selected functional groups (e.g., legs) while preserving the rest.
    Extended generation, editing, and mesh-animation results are provided as supplementary videos.\vspace{-4mm}}
  \label{fig:teaser}
\end{figure}

Modeling motion for articulated objects of arbitrary skeleton topology (humans, quadrupeds, birds, insects, mechanical rigs, and stylized 3D assets) remains a difficult problem. Recent text-to-motion generators~\cite{petrovich2023tmr,guo2026snapmogen} achieve impressive fidelity on HumanML3D~\cite{guo2022generating} and related human benchmarks~\cite{athanasiou2024motionfix}, but their motion representations are tied to a fixed SMPL human skeleton~\cite{SMPL} with fixed joint count, ordering, and connectivity. Extensions to animal categories have been proposed~\cite{jakab2024farm3d,yao2022lassie,wu2023dove,sun2024ponymation,li2024learning}, but they likewise assume a predefined skeleton topology one way or the other. Cross-topology motion generators have recently surfaced: SinMDM~\cite{raab2023single} trains a separate model from a single motion sequence, while AnyTop~\cite{gat2025anytop} uses a skeleton-conditioned diffusion model that operates at the per-joint level. These methods demonstrate important progress, but they do not provide a fixed-size discrete motion vocabulary shared across rigs. As a result, directly reusing a learned motion codebook for heterogeneous source-to-target motion transfer or token-based generation remains difficult across characters with different joint counts and connectivity.

Arguing that the appropriate abstraction level for cross-topology motion is neither the whole body nor individual joints, but functional groups of joints, we introduce \textbf{SAMoR}, a \textbf{s}keleton-\textbf{a}ware cross-topology \textbf{m}otion \textbf{r}epresentation for articulated objects. %, a cross-topology motion VQ-VAE for arbitrary skeletons.
SAMoR uses a graph-transformer encoder to model skeleton structure, cross-attention pooling
to compress heterogeneous per-joint motion features into $K=8$ part tokens, and
residual vector quantization to form a shared discrete motion codebook \cite{van2017neural}. A
target-skeleton-conditioned decoder reconstructs motion on the same or a
different rig, enabling cross-topology motion transfer directly at decode time.
The key challenge is query collapse: without explicit guidance, all $K$ queries attend broadly and produce redundant representations, reducing $K{=}8$ to the effective equivalent of $K{=}1$; we address this with an attention supervision loss that anchors each query toward a distinct functional joint group.
We train and evaluate SAMoR on a unified heterogeneous corpus curated from
HumanML3D~\cite{guo2022generating}, Truebones Zoo~\cite{truebones2022}, and
animated Objaverse-XL~\cite{deitke2023objaverse} assets, with held-out characters used
to test generalization to unseen skeletons. Importantly, when instantiated in a
HumanML3D-compatible setting, SAMoR remains competitive with fixed-skeleton
specialists on standard HumanML3D VQ-VAE reconstruction and text-to-motion
benchmarks. Yet unlike these specialists, the same part-token architecture
extends naturally to heterogeneous skeletons in our cross-topology setting,
without requiring per-character models.

%Concurrent with our work, MoCapAnythingV2 studies arbitrary-skeleton motion capture from monocular video through end-to-end pose-to-rotation learning. In contrast, SAMoR focuses on learning a shared discrete motion representation and generative token space across heterogeneous skeleton topologies.

Overall, our contributions are:
\begin{itemize}[nosep,leftmargin=*]
    \item \textbf{Graph-Transformer-based Motion Tokenization:}
    A cross-topology motion VQ-VAE that compresses
    arbitrary-skeleton motion into $K{=}8$ shared part tokens via
    graph-transformer encoding and cross-attention pooling, yielding a
    single discrete motion vocabulary across heterogeneous skeletons with varying joint
    counts and connectivity.

    \item \textbf{Attention Supervision for Query Specialisation:}
    An attention supervision loss that breaks query symmetry by seeding
    each query toward a distinct functional joint group, preventing
    collapse. Combined with joint-name dropout, this generalises to
    unseen skeletons with missing or non-canonical joint names.

    \item \textbf{Unified Heterogeneous Motion Corpus:}
    We curate a cross-topology corpus of ${\sim}71$k characters and ${\sim}100$k motions from 1 SMPL skeleton (HumanML3D), 65 animal species (Truebones Zoo), and ${\sim}71$k animated Objaverse-XL characters, with automated canonical joint-matching, character-level train/val splits,
    functional-group annotations across diverse morphologies, and per-character motion captions generated via video rendering and vision-language captioning~\cite{team2024gemini}. % to support text-conditioned generation.

    \item \textbf{Downstream Animation Tasks:}
    We show that the shared part-token space supports cross-topology
    motion transfer, text-conditioned generation via MaskGIT~\cite{chang2022maskgit}, and localized part-wise token editing.
\end{itemize}
%\begin{itemize}[nosep,leftmargin=*]
 %   \item A topology-agnostic discrete motion representation based on learned semantic part tokens shared across heterogeneous skeletons
 %   \item An attention-supervised part decomposition mechanism that prevents query collapse and enables consistent semantic routing across arbitrary rigs.
 %   \item Downstream applications in unified cross-topology reconstruction, transfer, text-conditioned generation, and localized editing within a shared token space.
%\end{itemize}
Extensive evaluations show that SAMoR substantially outperforms both single-token global representations ($K=1$) and per-joint tokenizations ($K=J$), validating intermediate-granularity part tokenization as an effective representation for cross-topology motion modeling.  We will make our implementation publicly available. 
\vspace{-1.5mm}
\section{Related Work}
\label{sec:related}
\vspace{-1.5mm}
\paragraph{Text-to-Motion Generation.}
Beyond unconditional pose~\cite{nadar2026posed,he2024nrdf} and motion~\cite{rempe2021humor,yu2026geometric} generation, text-driven human motion generation has advanced rapidly on benchmarks
such as HumanML3D~\cite{guo2022generating} and
KIT-ML~\cite{plappert2016kit}. Existing methods include
diffusion-based generators~\cite{tevet2022human,chen2023executing},
autoregressive discrete-token models~\cite{zhang2023generating,jiang2023motiongpt},
and masked-token generators~\cite{guo2024momask,pinyoanuntapong2024mmm,
xiao2025motionstreamer,yuan2024mogents,he2026molingo}, which achieve strong
motion quality, but their motion representations are tied to a fixed
human skeleton, typically SMPL~\cite{SMPL} with a fixed joint ordering and
connectivity. In contrast, SAMoR studies the orthogonal problem of
learning a shared discrete motion representation across heterogeneous
skeleton topologies.

\paragraph{Cross-Topology Motion Learning.}
Cross-skeleton retargeting~\cite{aberman2020skeleton} and cross-morphology alignment~\cite{li2024walkthedog} enable topology-flexible motion transfer on specific source--target pairs but do not yield a shared representation reusable across arbitrary rigs.
Part-based articulated-object methods such as LEPARD~\cite{liu2023lepard} study part discovery for 3D articulated shapes rather than motion generation.
On the generative side, SinMDM~\cite{raab2023single} trains a dedicated model per character.
AnyTop~\cite{gat2025anytop} handles heterogeneous skeletons at the per-joint level ($K{=}J$), preventing vocabulary sharing across different joint counts.
Concurrent with our work, NECromancer~\cite{xu2026necromancer} adopts a single-token ($K{=}1$) formulation.
SAMoR takes an intermediate approach: $K{=}8$ learned tokens via cross-attention pooling yield a shared codebook that neither $K{=}1$ nor $K{=}J$ supports natively.

\paragraph{Part-Based Motion Representations.}
Part-aware methods for human motion synthesis include per-part
tokenizers~\cite{zou2024parco}, part-level attention~\cite{zhong2023attt2m},
text-to-part conditioning~\cite{wang2025fg}, and compositional
generation~\cite{athanasiou2023sinc}.
All assume a fixed human skeleton with hard-coded part definitions.
SAMoR instead learns $K$ functional clusters of joints that generalise
across heterogeneous skeletons without requiring hand-defined
per-skeleton part assignments at inference time.

\paragraph{Automatic Rigging and Skinning.}
Automatic rigging converts raw 3D meshes into animation-ready characters.
Classical approaches such as Pinocchio~\cite{baran2007automatic} embed a
predefined skeleton template via geometric fitting.
Recent learned methods predict skeleton and skinning weights directly from
mesh geometry: RigNet~\cite{xu2020rignet} uses a graph neural network,
while MagicArticulate~\cite{song2025magicarticulate},
AnyMate~\cite{deng2025anymate}, and
Puppeteer~\cite{song2025puppeteer} extend this to end-to-end skeleton and
skinning prediction for diverse 3D objects.
SAMoR is complementary to these systems: given a rigged skeleton from any
such frontend, SAMoR provides the motion generation and transfer layer.

%Remove background session
% \input{sec/3_background}
\vspace{-1.5mm}
\section{Method}
\label{sec:method}
\begin{figure}[t]
    \centerline{\includegraphics[width=\dimexpr\linewidth]{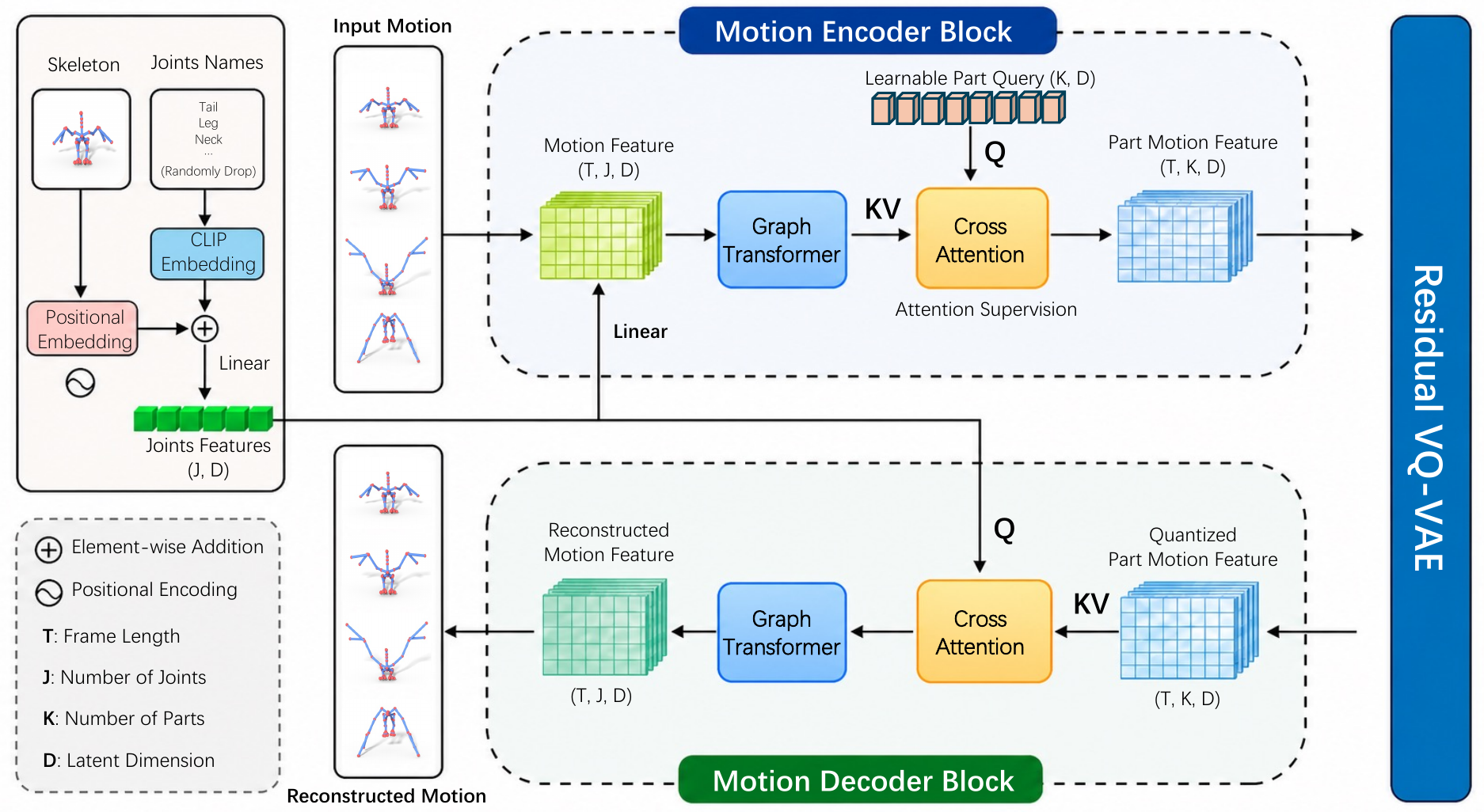}}
    \caption{
    \textbf{SAMoR architecture.}
    Per-joint motion features $(T,J,D)$ are fused with skeleton conditioning
    (pose anchor, joint-name embeddings, and graph biases)
    by a graph transformer. $K=8$ part queries pool the joints via
    attention-supervised cross-attention into part tokens $(T,K,D)$, which are
    quantized by residual VQ. The decoder mirrors the encoder with inverse
    cross-attention conditioned on the target skeleton, enabling
    cross-topology motion transfer at decode time. 
    % \TB{Pleaes render these skeletons as you render the other ones in experiments and use a better BG color? or different BG color to indicate different things?}
    }
    \label{fig:arch}
    \vspace{-4mm}
\end{figure}

\vspace{-1.5mm}
\paragraph{Overview.}
SAMoR is a cross-topology motion representation: each motion chunk on an
arbitrary skeleton is encoded into a discrete token sequence factorized over
$K=8$ part-level tokens at every downsampled temporal position
(\cref{fig:arch}). A skeleton-aware part VQ-VAE pools per-joint features
into $K$ part queries via cross-attention and quantizes them with residual VQ.
A target-skeleton-conditioned decoder expands the part tokens back to any rig,
enabling reconstruction and cross-topology transfer.

\paragraph{Topology representation.}
We represent a skeleton as a \emph{rooted kinematic tree} with $J$ joints, parent
map $\mathrm{pa}(j)$, and a per-joint pose anchor
$\{\mathbf a_j \in \mathbb{R}^3\}_{j=0}^{J-1}$ describing the target joint
geometry used for conditioning. Rooted kinematic
trees cover virtually all character animation rigs; the graph
structural biases (kinematic-tree hop distance, chain membership, edge
type) are defined for tree structures, though the transformer
architecture itself does not assume a tree. In the cross-topology setting,
$\mathbf a_j$ is the first-frame joint location; in the
HumanML3D-compatible setting, where a canonical SMPL rest pose is available,
it is the rest-pose joint location. A motion chunk of length $T$ is
represented as $\mathbf X \in \mathbb{R}^{T \times J \times d}$, where
$d=12$ in the full-feature setting (position, velocity, and 6D rotation per
joint) and $d=6$ when rotation is omitted. SAMoR encodes this chunk into a
discrete token tensor
$\mathbf Z \in \{1,\ldots,V\}^{(T/4)\times K \times R}$, with $T/4$
temporal positions after $4\times$ downsampling, $K=8$ part queries, and
$R$ residual quantization levels from a shared codebook of size $V$.

\vspace{-2mm}
\subsection{Skeleton-Aware Part VQ-VAE}
\label{sec:method:vqvae}
\vspace{-2mm}
\paragraph{Topology-agnostic motion input.}
For a joint $j$ at frame $t$, the full-feature representation is
$\mathbf x_j^t = [\mathbf p_j^t, \mathbf v_j^t, \mathbf R_j^t] \in
\mathbb{R}^{12}$, where $\mathbf p_j^t$ and $\mathbf v_j^t$ are the joint
position and velocity, and $\mathbf R_j^t$ is the local joint rotation in
the 6D continuous representation~\cite{zhou2019continuity}. In the
cross-topology setting, local rotations are defined relative to
rig-specific rest poses and are not directly comparable across arbitrary
skeletons. We therefore use root-normalized joint positions and velocities,
$\mathbf x_j^t=[\mathbf p_j^t,\mathbf v_j^t]\in\mathbb{R}^{6}$, as the
shared cross-rig signal. Global root translation is excluded because
heterogeneous assets have inconsistent reference frames and rest-pose
definitions; for HumanML3D, where the standard evaluator expects the 263-D
representation, root channels are modelled by a separate lightweight VQ
branch (\cref{sec:exp:humanml3d}).

\paragraph{Skeleton conditioning with pose anchors.}
We condition the encoder and decoder on the target skeleton through three
signals. First, a per-joint \emph{pose-anchor embedding} maps anchor
locations $\mathbf a_j$ to the model dimension through an MLP and re-injects
them at every graph-transformer layer. For cross-topology settings we use the
first frame as the anchor (more reliable than asset-provided rest poses for
heterogeneous sources such as OXL); for HumanML3D we use the canonical SMPL
rest-pose anchor. The pose anchor is a static skeleton-geometry signal only:
at transfer or generation time it can be any available posed frame or reference
pose of the target rig and is not required to correspond to the source or
target motion. Second, we add CLIP~\cite{radford2021learning} embeddings of
raw joint names, projected to the model dimension. Third, we inject graph
structure by adding learned biases to spatial attention logits: kinematic-tree
hop distance, same-chain indicator, and edge type.

\paragraph{Graph-transformer encoder.}
The encoder is a stack of $L$ transformer blocks, each alternating temporal
self-attention along $T$ for each joint with spatial self-attention along $J$
for each frame. Spatial attention uses the graph structural biases above;
we refer to these blocks as \emph{graph-transformer blocks} throughout.
A strided 1D convolution downsamples the temporal axis by $4\times$,
producing $\mathbf H\in\mathbb{R}^{B\times (T/4)\times J\times D}$.
Full architectural details and hyperparameters are in \cref{app:impl}.

\paragraph{Part pooling: joints to part tokens.}
$K$ learnable query vectors compress the $J$ encoded joints into $K$
fixed-size part tokens via cross-attention, followed by a residual connection,
layer normalization, and a $K{\times}K$ self-attention layer for part-to-part
communication.
This operation is agnostic to skeleton identity: $J$ is variable and the
queries impose no constraints on which joints are present---the same
architecture processes a 5-joint prop, a 22-joint humanoid, and a 96-joint
creature without modification.

Without supervision, the $K$ queries have no incentive to specialize and
collapse toward near-identical attention distributions, making $K{=}8$
behave no better than $K{=}1$ in practice (\cref{sec:method:attn}).
To break this symmetry, we supervise the cross-attention maps during training
using canonical functional-group labels $y_j\in\{0,\ldots,K{-}1\}\cup\{-1\}$ derived
automatically from joint names.
Crucially, joints that cannot be matched to any label receive $y_j{=}{-1}$
and contribute \emph{zero} terms to the supervision loss; they are still
encoded and routed freely by the learned attention.
For skeletons with no canonical joint names at all (mechanical rigs,
procedural assets), every joint has $y_j{=}{-1}$ and the $K$ slots become
purely unsupervised functional buckets, driven by kinematic structure and
motion patterns alone.

We set $K{=}8$ to identify eight functional clusters that, as abstract
skeleton semantics, encapsulate the major motion-producing structures
across the diverse topologies in our training corpus.
For readability, we label these clusters as
\emph{lower torso}, \emph{upper torso}, \emph{head/neck},
\emph{left arm} (forelimb/wing), \emph{right arm},
\emph{left leg} (hindlimb), \emph{right leg}, and \emph{tail} ---
though these captions are interpretive handles rather than strict
anatomical requirements: a cluster labelled `arm' may capture insect
appendages, mechanical links, or wing joints depending on the input
skeleton.
The attention map $\boldsymbol{\alpha}\in[0,1]^{K\times J}$ is the target
of the supervision loss in \cref{sec:method:attn}.
Inference-time part-assignment visualisations for diverse topologies,
including non-biological rigs, are in \cref{app:qual}.

\paragraph{Quantization, decoder, and training objective.}
Each part token is quantized by an $R$-level residual VQ module with a shared
codebook; quantize-dropout randomly drops deeper RVQ levels during training to
improve codebook utilization. The decoder mirrors the encoder: it expands the
$K$ quantized part tokens back to $J$ target joints through inverse
cross-attention, where target-joint queries are conditioned on the target
skeleton's pose anchor, joint-name embeddings, and graph biases. Since the
decoder is conditioned on the \emph{target} skeleton, the same source tokens
can be decoded on a different rig for cross-topology transfer without
retraining. The VQ-VAE is trained with smooth-$\ell_1$ position and velocity
reconstruction losses, a VQ commitment loss, a bone-length consistency loss
(squared parent-to-child distance error, ablated in \cref{tab:oxl_abl}), and
the attention supervision loss $\mathcal{L}_\text{attn}$
(\cref{sec:method:attn}). A rotation reconstruction loss is added in the
HumanML3D-compatible setting.

\vspace{-2mm}
\subsection{Attention Supervision for Query Specialisation}
\label{sec:method:attn}
\vspace{-2mm}
\paragraph{Part-query collapse in naive training.}
Given per-frame joint features
$\mathbf X=(\mathbf x_1,\ldots,\mathbf x_J)\in\mathbb{R}^{d\times J}$ and
$K$ learnable part queries $\{\mathbf q_k\}_{k=1}^K\subset\mathbb{R}^d$,
the part-pool cross-attention computes each part token:
\begin{equation}
\mathbf p_k =
\mathrm{LN}\left(
\mathbf q_k + \sum_{j=1}^{J} \alpha_{k,j} W_V \mathbf x_j
\right),
\qquad
\alpha_{k,j} =
\frac{
\exp\left((W_Q\mathbf q_k)^\top(W_K\mathbf x_j)/\sqrt{d}\right)
}{
\sum_{j'=1}^{J}
\exp\left((W_Q\mathbf q_k)^\top(W_K\mathbf x_{j'})/\sqrt{d}\right)
}.
\label{eq:part_pool}
\end{equation}
Without supervision on $\boldsymbol{\alpha}$, the residual connection lets
queries differentiate via their learned embeddings alone, with no incentive
for $\alpha_{k,j}$ to differ across $k$. Empirically, attention maps drift
toward near-uniform distributions, and reconstruction quality at $K=8$ equals
that at $K=1$ (\cref{tab:oxl_abl}).

\paragraph{Attention supervision loss.}
We supervise $\boldsymbol{\alpha}$ directly with per-joint functional-group labels
$y_j\in\{0,\ldots,K{-}1\}\cup\{-1\}$, where $-1$ marks unknown joints using the row-wise cross-entropy: 
\begin{equation}
\mathcal L_{\mathrm{attn}}
=
-\sum_{k:|S_k|>0}
\sum_{j=1}^{J}
T_{k,j}\log \alpha_{k,j}, \quad\text{where}\quad
T_{k,j} =
\begin{cases}
1/|S_k|, & \text{if } y_j=k,\\
0, & \text{otherwise}
\end{cases}
\label{eq:attn_loss}
\end{equation}
$S_k=\{j:y_j=k\}$ and $T$ is the uniform target distribution for each non-empty part $k$. 
% Let $S_k=\{j:y_j=k\}$; for each non-empty part $k$ we define the uniform target distribution
% \begin{equation}
% T_{k,j} =
% \begin{cases}
% 1/|S_k|, & \text{if } y_j=k,\\
% 0, & \text{otherwise},
% \end{cases}
% \label{eq:attn_target}
% \end{equation}
% and minimize the row-wise cross-entropy
%\begin{equation}
%\mathcal L_{\mathrm{attn}}
%=
%-\sum_{k:|S_k|>0}
%\sum_{j=1}^{J}
%T_{k,j}\log \alpha_{k,j}.
%\label{eq:attn_loss}
%\end{equation}
Unlabeled joints ($y_j{=}{-1}$) contribute no terms; labels are obtained
automatically from joint names via canonical matching (\cref{app:matching}).
Labels are used only during training;
at inference, joints are routed by the learned cross-attention alone.
Gradients flow through the softmax to $W_Q$, $W_K$, and the encoder features,
encouraging functional cluster specialisation. \cref{eq:part_pool,eq:attn_loss}
are written per temporal position; in practice the loss is averaged over batch
and time.

\paragraph{Joint-name dropout.}
Joint names enter the encoder as CLIP embeddings. We randomly replace each
embedding with a shared zero vector during training; since
$\mathcal L_{\mathrm{attn}}$ is still applied, the model learns to route
joints from motion patterns and skeleton topology rather than name strings
alone, improving robustness to unseen rigs with missing or non-semantic names.

\vspace{-1.5mm}
\subsection{Token Generation, Editing, and Mesh Animation}
\label{sec:method:generation}
\vspace{-1.5mm}
\paragraph{Text-conditioned token generation.}
We train a MaskGIT~\cite{chang2022maskgit} transformer over the $(T/4)\times K$ grid of SAMoR tokens, conditioned on text and the target skeleton. The main transformer predicts
level-0 RVQ code indices, and a lightweight residual transformer predicts the
remaining RVQ levels conditioned on previously decoded codes, following
masked-token motion generators. Since generation happens in SAMoR's shared
part-token space, the predicted tokens can be decoded on arbitrary target
skeletons through the same target-skeleton-conditioned decoder.

\paragraph{Part-wise editing and mesh animation.}
For editing, we encode a motion into SAMoR tokens and mask a selected
subset of part slots; the generator in-fills only the masked slots,
localizing changes to the selected functional groups.
For raw meshes, any off-the-shelf auto-rigging
system~\cite{song2025puppeteer,song2025magicarticulate,deng2025anymate}
provides the skeleton and skinning weights; SAMoR then generates or
transfers motion, recovering local rotations via IK before applying LBS.

\section{Experiments}
\label{sec:exp}

\begin{figure}[!t]
  \centerline{\includegraphics[width=\linewidth]{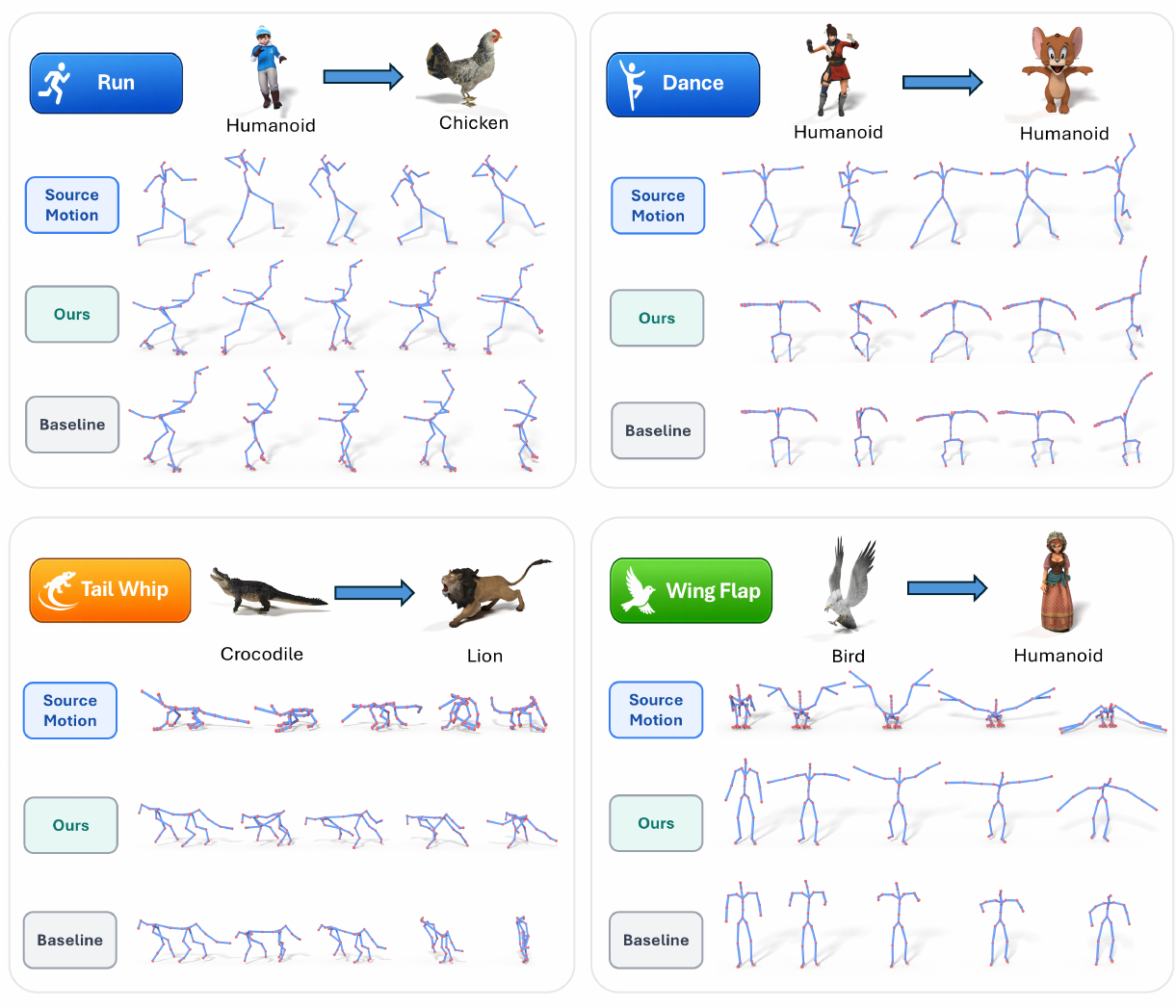}}
  \caption{\textbf{Cross-topology motion transfer.} Each $2{\times}2$
    panel shows \emph{Source Motion}, \emph{SAMoR} ($K{=}8$), and the
    single-query \emph{Baseline} ($K{=}1$, same cross-topology decoder).
    SAMoR preserves appendage dynamics across topology shifts;
    $K{=}1$ retains coarse body rhythm but loses limb, tail, and wing
    detail, consistent with the reconstruction and round-trip gaps in \cref{tab:oxl_abl}.
    All visualizations are root-relative.}
  \label{fig:transfer_qual}
  \vspace{-3mm}
\end{figure}

% -----------------------------------------------------------------------
\subsection{Experimental Setup}
\label{sec:exp:setup}
% -----------------------------------------------------------------------
\paragraph{Datasets.}
We use three datasets covering the cross-topology spectrum.
\emph{HumanML3D (H3D)}~\cite{guo2022generating} provides $\sim$23K SMPL-22
humanoid motion-caption pairs in the standard 263-D feature format
expected by the MoMask evaluator (root trajectory, relative joint
positions, 6D joint rotations, joint velocities, and foot-contact
indicators concatenated per frame).
\emph{Truebones Zoo}~\cite{truebones2022} is a motion-capture library of 65
animal species (quadrupeds, birds, reptiles, insects) with varying joint
counts and named motion clips.
\emph{Objaverse-XL (OXL)}~\cite{deitke2023objaverse} is our curated subset of rigged
and skinned animated characters from Objaverse-XL spanning humanoid, quadruped, and non-biological rigs.
All motions are converted to a unified format of per-joint positions
$\mathbf{p}_j^t$ and velocities $\mathbf{v}_j^t$; for OXL and Zoo, the first frame of each motion serves as the pose
anchor; for HumanML3D-compatible experiments, the canonical SMPL rest
pose is used. Character-level
splits guarantee that OXL validation characters are never seen
during training. For the optional text-conditioned generator, one motion caption per OXL
character is generated by rendering the skeleton motion sequence and
textured mesh sequence and captioning the resulting video with a
vision-language model (Gemini Flash~\cite{team2024gemini}); these captions
are only used to train the SAMoR MaskGIT generation model.

All characters are normalized by mesh bounding box, up-axis, and facing
direction. During training, $T{=}64$ frames are randomly sampled from
each motion clip; full preprocessing details are in \cref{app:data}.
\cref{tab:dataset_stats} summarises dataset scale.

\begin{table}[h]
  \centering
  \caption{Dataset statistics. During training, $T{=}64$ frames are randomly
    sampled from each motion clip at 20\,fps.
    OXL and Zoo splits are character-disjoint; H3D uses the standard split
    protocol with a single shared SMPL skeleton.
    Joint counts are the range across the corpus.}
  \label{tab:dataset_stats}
  \small
  \begin{tabular}{@{}l r r r r r@{}}
    \toprule
    Dataset & Train entities & Val entities & Train motions & Val motions & Joints \\
    \midrule
    HumanML3D~\cite{guo2022generating}
      & 1 (SMPL) & 1 (SMPL) & 23,384 & 4,384 & 22 \\
    Truebones Zoo~\cite{truebones2022}
      & 54 species & 11 species & 939 & 110 & 9--71 \\
    Objaverse-XL (ours)
      & 67,646 & 3,528 & 67,646 & 3,528 & 5--96 \\
    \midrule
    \textbf{Total}
      & 67,701 & 3,540 & 91,969 & 8,022 & 5--96 \\
    \bottomrule
  \end{tabular}
\end{table}

\paragraph{Baselines.}
For HumanML3D (\cref{tab:humanml3d}) we compare against
T2M-GPT~\cite{zhang2023generating}, MoMask~\cite{guo2024momask},
and MoGenTS~\cite{yuan2024mogents} for both text-to-motion
generation and VQ-VAE reconstruction. For cross-topology
reconstruction (\cref{tab:oxl_abl}) we adapt both tokenizers to
variable-$J$ skeletons: T2M-GPT is extended with a single-stream
wrapper that treats all $J$ joints as a flat per-frame feature
($K{=}1$ style), while MoGenTS is extended to process per-joint
position-and-velocity features for arbitrary $J$, producing a
$T{\times}J$ grid of per-joint tokens ($K{=}J$ style). Both are
padded to the max joint count ($J{=}96$); padded positions are
masked during attention and loss computation for fairness.
These adaptations instantiate two natural tokenizer extremes---global
fixed-size tokenization and per-joint tokenization---and are not
intended as optimised arbitrary-topology systems.
Motion generators such as AnyTop~\cite{gat2025anytop} synthesize new
motions for a given skeleton topology rather than encoding an existing
motion into a reusable discrete latent, and therefore do not
participate in the reconstruction or transfer evaluations.

\paragraph{Metrics.}
On HumanML3D we report FID, R-Precision at Top-1/2/3, MM-Dist,
and MPJPE (mm) for VQ-VAE reconstruction. The MPJPE convention is
computed on the 22-joint relative-pose representation in the
canonical body frame (root excluded); SAMoR-unified enters the same
column under the body-joint-only protocol (global root and
foot-contact channels excluded) after inverting bounding-box
normalization, and is not evaluated under the full 263-D protocol.
For cross-topology reconstruction we report MPJPE in $10^{-2}$
normalized units (bounding-box frame, per-frame root subtracted)
and bone-length error.
Unless otherwise stated, all cross-topology experiments use the
shared $d{=}6$ position-and-velocity representation; local rotations
are used only in HumanML3D-compatible settings where a consistent
rest pose is available.
Metric definitions are in \cref{app:metrics}.

% -----------------------------------------------------------------------
\vspace{-1.5mm}
\subsection{HumanML3D Benchmark}
\label{sec:exp:humanml3d}
% -----------------------------------------------------------------------
\vspace{-1.5mm}
To handle the root trajectory and foot-contact channels in HumanML3D's
263-D representation, we add a parallel 1D-convolutional VQ branch for
global channels following MoGenTS~\cite{yuan2024mogents}.
\cref{tab:humanml3d} reports both text-to-motion generation (top) and
VQ-VAE reconstruction (bottom).

\paragraph{Results.}
On generation, SAMoR-H3D's FID of $0.039$ sits between MoGenTS
($0.033$) and MoMask ($0.045$); Top-3 R-Precision ties MoGenTS at
$0.812$ and MM-Dist ($2.897$) is competitive, confirming that the
cross-topology part tokenizer does not hurt fixed-skeleton generation
quality. On reconstruction, SAMoR-H3D ($17.3$ mm MPJPE, $0.008$ FID)
beats MoMask and T2M-GPT by large margins and leads MoGenTS on
Top-1, Top-3, and MM-Dist despite higher MPJPE. SAMoR-unified is evaluated under the root-excluded body-joint protocol
only (no root or foot-contact channels) and does not participate in the
full 263-D metrics (FID, R-Precision, MM-Dist); its body-joint MPJPE of
$15.1$ mm is comparable within this root-excluded MPJPE protocol, indicating that
heterogeneous training preserves a useful humanoid body-joint sub-vocabulary.

% --- Table 1: HumanML3D ---
\begin{table}[t]
  \centering
  \caption{HumanML3D benchmark. Top: text-to-motion generation;
    bottom: VQ-VAE reconstruction. MPJPE in mm (22-joint
    canonical-frame, root excluded). $\dagger$~numbers from papers.}
  \label{tab:humanml3d}
  \small\setlength{\tabcolsep}{3pt}
  \begin{tabular}{@{}l c ccc c c@{}}
    \toprule
    Method & FID$\downarrow$ & Top1$\uparrow$ & Top2$\uparrow$ & Top3$\uparrow$
           & MM-Dist$\downarrow$ & MPJPE$\downarrow$ \\
    \midrule
    Real motion
      & 0.002\tiny{$\pm$.000}
      & 0.511\tiny{$\pm$.003} & 0.703\tiny{$\pm$.003} & 0.797\tiny{$\pm$.002}
      & 2.974\tiny{$\pm$.008} & — \\
    \midrule
    \multicolumn{7}{@{}l}{\textit{Text-to-motion generation}} \\
    \quad T2M-GPT$^\dagger$
      & 0.141\tiny{$\pm$.005}
      & 0.492\tiny{$\pm$.003} & 0.679\tiny{$\pm$.002} & 0.775\tiny{$\pm$.002}
      & 3.121\tiny{$\pm$.009} & — \\
    \quad MoMask$^\dagger$
      & 0.045\tiny{$\pm$.002}
      & 0.521\tiny{$\pm$.002} & 0.713\tiny{$\pm$.002} & 0.807\tiny{$\pm$.002}
      & 2.958\tiny{$\pm$.008} & — \\
    \quad MoGenTS$^\dagger$
      & \textbf{0.033}\tiny{$\pm$.001}
      & \textbf{0.529}\tiny{$\pm$.003} & \textbf{0.719}\tiny{$\pm$.002} & \textbf{0.812}\tiny{$\pm$.002}
      & \textbf{2.867}\tiny{$\pm$.006} & — \\
    \quad \textbf{SAMoR-H3D}
      & 0.039\tiny{$\pm$.002}
      & 0.523\tiny{$\pm$.002} & 0.715\tiny{$\pm$.003} & \textbf{0.812}\tiny{$\pm$.001}
      & 2.897\tiny{$\pm$.008} & — \\
    \midrule
    \multicolumn{7}{@{}l}{\textit{VQ-VAE reconstruction}} \\
    \quad T2M-GPT$^\dagger$
      & 0.070 & 0.489 & 0.676 & 0.775 & 3.130 & 58.0 \\
    \quad MoMask$^\dagger$
      & 0.019\tiny{$\pm$.000}
      & 0.508\tiny{$\pm$.003} & 0.701\tiny{$\pm$.002} & 0.795\tiny{$\pm$.002}
      & 2.999\tiny{$\pm$.006} & 29.5 \\
    \quad MoGenTS$^\dagger$
      & \textbf{0.005}\tiny{$\pm$.000}
      & 0.504\tiny{$\pm$.002} & \textbf{0.702}\tiny{$\pm$.002} & 0.797\tiny{$\pm$.002}
      & 2.978\tiny{$\pm$.006} & \textbf{13.8} \\
    \quad \textbf{SAMoR-H3D}
      & 0.008\tiny{$\pm$.000}
      & \textbf{0.514}\tiny{$\pm$.001} & 0.701\tiny{$\pm$.002} & \textbf{0.798}\tiny{$\pm$.002}
      & \textbf{2.977}\tiny{$\pm$.005} & 17.3 \\
    \midrule
    \multicolumn{7}{@{}l}{\textit{Cross-topology model, evaluated on H3D body joints}} \\
    \quad \textbf{SAMoR-unified}$^\ddagger$ {\small(body-joint only)}
      & — & — & — & — & — & 15.1 \\
    \bottomrule
  \end{tabular}
  \\[0.3em]
  {\footnotesize
    $^\ddagger$Trained on the full cross-topology corpus; MPJPE is
    computed under the same root-excluded 22-joint body protocol as
    the reconstruction baselines and is comparable within this
    root-excluded MPJPE protocol. Full 263-D metrics (FID, R-Precision, MM-Dist)
    involving root and foot-contact channels are not reported. %For our model, we use body-joint MPJPE only.
    }
\vspace{-0.9em}
\end{table}

% -----------------------------------------------------------------------
\vspace{-2mm}
\subsection{Cross-Topology Reconstruction and Ablations}
\label{sec:exp:crosstopo}
% -----------------------------------------------------------------------
\vspace{-2mm}
\paragraph{Setup and baselines.}
We evaluate on the curated SAMoR validation set: HumanML3D humanoid clips,
held-out OXL characters (\emph{zero character overlap} with training),
and Truebones Zoo animals. Per-source columns in \cref{tab:oxl_abl}
isolate each regime. Baseline adaptations are as described in \cref{sec:exp:setup};
the strongest, MoGenTS (per-joint adapt., $K{=}J$), reaches
$16.14\!\times\!10^{-2}$ MPJPE, $5.8\!\times$ above SAMoR.

\paragraph{Per-source analysis.}
SAMoR leads on every source; the gap is largest on H3D
($2.28$ vs.\ $16.02$, $7.0\!\times$). Ablation rows trace the gains
to structural inductive biases: K=1 raises Zoo MPJPE by $+82\%$;
removing $\mathcal{L}_\text{attn}$ nearly doubles the average error
and consistently degrades every source, while also raising Bone-Error
from $0.79$ to $1.16$, showing the model loses skeleton fidelity,
not just positional accuracy.
RVQ provides consistent gains across sources; removing it raises
source-macro MPJPE by $44\%$.

\paragraph{Ablation summary.}
The ablations confirm three contribution tiers.
\emph{Core:} $K{>}1$ queries with $\mathcal{L}_\text{attn}$ are
essential---removing either raises MPJPE by $83{-}100\%$.
Crucially, $K{=}8$ without supervision ($5.04$) nearly matches
$K{=}1$ ($5.49$), confirming that supervision---not query count---drives
structured routing; we therefore ablate these two alternatives to isolate
query count from cluster specialisation.
\emph{Structural} components each contribute $8{-}10\%$, with RVQ
accounting for $+44\%$ when removed. %---placing residual coding closer to the core tier.
\emph{Name conditioning} costs only $13{-}16\%$ when omitted, keeping
the pipeline robust to non-semantic rigs; this moderate impact is
expected given that canonical label coverage is $100\%$ on H3D but
only $69\%$ on OXL and $81\%$ on Zoo (details in \cref{app:name_dropout}).

\paragraph{Joint-name conditioning.}
Both name-ablation rows stay within $+16\%$ of the full model,
confirming SAMoR does not rely on joint-name strings as a shortcut.
The full dropout-probability sweep is in \cref{app:name_dropout}.
Part-query assignment visualizations in \cref{app:qual} show that appendages such as insect legs, mechanical
limbs, and bird wings are routed to functional arm / leg slots even
without joint-name embeddings, indicating that skeleton topology
and motion patterns alone recover functionally meaningful routing.

% --- Table 3: OXL cross-topology reconstruction + ablations + RT ---
\begin{table}[t]
  \centering
  \caption{Cross-topology reconstruction and round-trip transfer consistency on the
    SAMoR validation set (character-disjoint from training).
    \textbf{MPJPE} and \textbf{BoneErr}: reconstruction quality per source split
    (H3D, OXL, Zoo); \emph{All} is the macro-average.
    \textbf{RT-MPJPE}: round-trip MPJPE—source$\!\to\!$canonical
    SMPL-22$\!\to\!$source (pelvis-aligned); reported for OXL and Zoo only
    (H3D shares the canonical SMPL-22 topology, so cross-topology
    transfer cost is not defined).
    $-$ indicates the method does not support cross-topology transfer.
    Values in $\times\!10^{-2}$ normalized units.}
  \label{tab:oxl_abl}
  \setlength{\tabcolsep}{3.2pt}
  \begin{tabular}{@{}l cccc c cc@{}}
    \toprule
    & \multicolumn{4}{c}{MPJPE\,$(\!\times\!10^{-2})\downarrow$}
      & BoneErr & \multicolumn{2}{c}{RT-MPJPE\,$(\!\times\!10^{-2})\downarrow$} \\
    \cmidrule(lr){2-5} \cmidrule(lr){7-8}
    Method / Variant & All & OXL & H3D & Zoo
      & $(\!\times\!10^{-2})\downarrow$ & OXL & Zoo \\
    \midrule
    \multicolumn{8}{@{}l}{\textit{Baselines (variable-$J$ adaptations)}} \\
    \quad T2M-GPT~\cite{zhang2023generating}
      & 35.77 & 35.03 & 36.59 & 35.64 & 11.76 & $-$ & $-$ \\
    \quad MoGenTS~\cite{yuan2024mogents} (per-joint adapt.)
      & 16.14 & 16.36 & 16.02 & 16.07 & 10.42 & $-$ & $-$ \\
    \midrule
    \multicolumn{8}{@{}l}{\textit{SAMoR ablations --- Architecture}} \\
    \quad $K{=}1$ (single part query)
      & 5.49 & 5.72 & 4.67 & 6.37 & 1.25 & 8.78 & 10.03 \\
    \quad w/o RVQ (single-level VQ)
      & 3.97 & 4.08 & 3.38 & 4.66 & 0.90 & 8.16 & 9.83 \\
    \quad w/o graph transformer
      & 3.03 & 2.93 & 2.56 & 3.69 & 0.88 & 7.53 & 9.65 \\
    \midrule
    \multicolumn{8}{@{}l}{\textit{SAMoR ablations --- Losses}} \\
    \quad w/o $\mathcal{L}_\text{attn}$ ($K{=}8$)
      & 5.04 & 4.67 & 4.53 & 5.98 & 1.16 & 8.38 & 9.89 \\
    \quad w/o bone-length loss
      & 2.97 & 2.62 & 2.29 & 3.44 & 0.91 & 6.97 & 9.67 \\
    \midrule
    \multicolumn{8}{@{}l}{\textit{SAMoR ablations --- Joint name conditioning}} \\
    \quad w/o joint-name emb.\ (train)
      & 3.10 & 2.97 & 2.50 & 3.86 & 0.85 & 5.87 & 9.48 \\
    \quad w/o joint-name emb.\ (infer.)
      & 3.19 & 2.73 & 2.69 & 4.06 & 0.84 & 6.06 & 9.27 \\
    \midrule
    \textbf{SAMoR (full)}
      & \textbf{2.75} & \textbf{2.46} & \textbf{2.28} & \textbf{3.50}
      & \textbf{0.79} & \textbf{5.74} & \textbf{8.72} \\
    \bottomrule
  \end{tabular}
\end{table}

% -----------------------------------------------------------------------
\vspace{-1.5mm}
\subsection{Cross-Topology Transfer and Round-Trip Consistency}
\label{sec:exp:transfer}
% -----------------------------------------------------------------------
\vspace{-1.5mm}
Cross-topology transfer does not have paired ground-truth target
motions: given a source motion decoded onto a different target skeleton, there is no unique correct target trajectory.
Hence, we evaluate transfer with (i) qualitative comparisons in \cref{fig:transfer_qual}, and (ii)
\emph{round-trip MPJPE} (RT-MPJPE) in \cref{tab:oxl_abl}, where a
motion is mapped source$\to$canonical SMPL-22$\to$source and
compared against the original.
RT-MPJPE is a transfer-consistency proxy;
\cref{fig:transfer_qual} provides the perceptual complement.

\paragraph{Transfer baseline and qualitative results.}
Fixed-skeleton tokenizers cannot decode to unseen skeleton topologies
(marked~``$-$'' in \cref{tab:oxl_abl}); we therefore use the
$K{=}1$ variant of SAMoR as the transfer baseline---the same
architecture and cross-topology decoder, but with a single part query
that collapses all joint features into one global token instead of
$K{=}8$ specialized part tokens.
\cref{fig:transfer_qual} shows SAMoR preserves appendage-specific
dynamics across topology shifts, while $K{=}1$ retains only coarse
body rhythm and loses limb, tail, and wing detail, consistent with
the reconstruction and round-trip gaps in \cref{tab:oxl_abl}.
Part-query
assignment visualizations are in \cref{app:qual}.

\paragraph{Round-trip consistency.}
The RT-MPJPE columns in \cref{tab:oxl_abl} provide a quantitative
proxy for transfer consistency, routing source motions through a
canonical SMPL-22 skeleton and back.
SAMoR achieves the lowest OXL RT-MPJPE ($5.74$) and Zoo RT-MPJPE
($8.72$) among all high-fidelity variants, indicating that the
part-token representation preserves source motion through an
intermediate topology.
$K{=}1$ and w/o\,$\mathcal{L}_\text{attn}$ substantially increase
RT-MPJPE ($8.78$/$8.38$ OXL; $10.03$/$9.89$ Zoo), showing that
collapsed or unsupervised pooling loses transfer-relevant part structure.
Removing RVQ or the graph transformer also degrades round-trip
consistency, suggesting that both discrete residual coding and
graph-aware encoding contribute to robust cross-topology decoding.

\vspace{-1.5mm}
\section{Conclusion}
\label{sec:conclusion}
\vspace{-1.5mm}
We presented SAMoR, a cross-topology motion VQ-VAE that encodes
arbitrary-skeleton motion into a fixed set of $K{=}8$ shared part-level
tokens. The central challenge is that naive multi-query pooling collapses
without supervision; our topology-agnostic attention supervision loss
addresses this by guiding each query toward a distinct functional joint cluster,
while joint-name dropout ensures robustness to unseen rigs with missing
or non-semantic joint names. Trained on a unified corpus of humanoid,
animal, and cross-topology characters, SAMoR reaches
$2.75\!\times\!10^{-2}$ normalized MPJPE---$5.8\!\times$ below the
strongest adapted variable-$J$ tokenizer baseline---while remaining competitive with
fixed-skeleton specialists on HumanML3D. The shared discrete codebook directly enables cross-topology transfer,
text-conditioned generation, and localized part-wise editing,
\ifdefined\arxivmode
with representative qualitative examples in the main paper, additional
qualitative results in \cref{app:qual}, and videos on our
\href{\projecturl}{project website}.
\else
with representative qualitative examples in the main paper and extended results as supplementary videos.
\fi

\paragraph{Limitations and future work.}
SAMoR operates on joint positions rather than rotations in the
cross-topology setting, because heterogeneous assets lack a unified
rest-pose reference that makes local rotations comparable across rigs.
Recovering bone rotations for LBS mesh animation therefore requires an
IK post-process, which introduces twist ambiguity; accurate twist
recovery from positions alone remains an open problem.
The $K{=}8$ functional grouping is also a fixed design choice;
learning the number and structure of groups automatically is a natural
direction for future work.

% The `ack` environment is automatically hidden in anonymous submission
% and shown in [final] / [preprint] mode.
\begin{ack}
% TODO: funding sources, compute, collaborators.
TB acknowledges support from the UKRI Engineering and Physical Sciences Research Council (EPSRC) through the Future Leaders Fellowship [grant number MR/Y018818/1].
GPM is a member of the Machine Learning Cluster of Excellence, EXC number 2064/1 -- Project number 390727645, and is endowed by the Carl Zeiss Foundation.
\end{ack}

%\section*{Acknowledgements}
%TB acknowledges support from the UKRI Engineering and Physical Sciences Research Council (EPSRC) through the Future Leaders Fellowship [grant number MR/Y018818/1]. 

% ---------------------------------------------------------------------
% Acknowledgments  
% ---------------------------------------------------------------------
% \input{sec/7_ack}

% ---------------------------------------------------------------------
% References
% ---------------------------------------------------------------------
% \clearpage  % flush all pending floats before references
%\bibliographystyle{abbrvnat}
\bibliographystyle{plain}
\bibliography{ref}

% ---------------------------------------------------------------------
% Appendix / Supplementary  (arXiv build: shared supp + extra figures)
% ---------------------------------------------------------------------
\appendix

\newpage
\appendix
\section*{Appendix}

% =====================================================================
% B. Implementation Details
% =====================================================================
\section{Implementation Details}
\label{app:impl}

\subsection{SAMoR Stage 1: Part VQ-VAE}

\paragraph{Architecture hyperparameters.}
\cref{tab:hyperparams} summarises the key architectural parameters.
The encoder and decoder each consist of $L{=}4$
\emph{ChainAwareGraphTransformerBlock} layers.  Each block alternates
temporal self-attention (standard multi-head attention along the $T$
axis per joint) and spatial graph-transformer attention (along the $J$
axis per frame) with a two-layer FFN.
Temporal downsampling by $4\times$ is achieved through two stacked
strided Conv1d layers (stride 2 each), producing $(T/4)$ temporal
positions.

\begin{table}[h]
\centering
\caption{SAMoR Stage 1 architecture and training hyperparameters.}
\label{tab:hyperparams}
\small
\begin{tabular}{lll}
\toprule
\textbf{Parameter} & \textbf{Value} & \textbf{Note} \\
\midrule
Model dimension $D$         & 512  & \\
Encoder/decoder blocks $L$  & 4 + 4 & \\
Attention heads             & 4    & head dim $= 128$ \\
FFN hidden dim              & 1024 & \\
Temporal downsampling       & $4\times$ & two stride-2 Conv1d \\
Max joints $J_{\max}$       & 96   & padded; real joints masked \\
Clip length $T$             & 64   & frames randomly sampled \\
Part queries $K$            & 8    & \\
RVQ depth $R$               & 6    & \\
Codebook size $V$           & 512  & per RVQ level, EMA updates \\
Quantize-dropout prob.      & 0.5  & drops deeper RVQ levels \\
Name-dropout prob.\ $p$     & 0.3  & \\
Max hop distance            & 20   & clamped for bias lookup \\
Edge types                  & 6    & see \cref{app:graph} \\
Dropout                     & 0.1  & \\
\midrule
Batch size                  & 64   & \\
Learning rate               & $10^{-4}$ & 5\% linear warmup \\
Optimizer                   & AdamW & $\beta_1{=}0.9$, $\beta_2{=}0.999$ \\
Precision                   & bf16  & \\
EMA decay                   & 0.99  & on codebook entries \\
Epochs                      & 100   & \\
Hardware                    & 4$\times$ GH200 96\,GB & \\
Training time               & $\sim$96\,h & wall-clock \\
\bottomrule
\end{tabular}
\end{table}

\paragraph{Training losses.}
The VQ-VAE is trained with the weighted sum:
\begin{equation}
\mathcal{L} =
  \lambda_{\text{pos}}\,\mathcal{L}_{\text{pos}}
+ \lambda_{\text{vel}}\,\mathcal{L}_{\text{vel}}
+ \lambda_{\text{vq}}\,\mathcal{L}_{\text{vq}}
+ \lambda_{\text{bone}}\,\mathcal{L}_{\text{bone}}
+ \lambda_{\text{smooth}}\,\mathcal{L}_{\text{smooth}}
+ \lambda_{\text{attn}}\,\mathcal{L}_{\text{attn}},
\label{eq:total_loss}
\end{equation}
where $\mathcal{L}_{\text{pos}}$ and $\mathcal{L}_{\text{vel}}$ are
smooth-$\ell_1$ losses on root-relative joint positions and
frame-to-frame velocities; $\mathcal{L}_{\text{vq}}$ is the VQ
commitment loss; $\mathcal{L}_{\text{bone}}$ is the bone-length
consistency loss (squared parent-to-child distance error);
$\mathcal{L}_{\text{smooth}}$ penalises temporal variation of the
quantized part-token latents
($\|z_q^{t+1}-z_q^{t}\|^2$ averaged over time); and
$\mathcal{L}_{\text{attn}}$ is the attention supervision loss.
Weights are given in \cref{tab:loss_weights}.
A rotation reconstruction loss is added only in the HumanML3D-compatible
setting.

\begin{table}[h]
\centering
\caption{Loss weights used during Stage 1 training.}
\label{tab:loss_weights}
\small
\begin{tabular}{lcl}
\toprule
Loss & Weight & Description \\
\midrule
$\mathcal{L}_{\text{pos}}$    & 1.0  & smooth-$\ell_1$ on joint positions \\
$\mathcal{L}_{\text{vel}}$    & 0.5  & smooth-$\ell_1$ on joint velocities \\
$\mathcal{L}_{\text{vq}}$     & 1.0  & VQ commitment + entropy \\
$\mathcal{L}_{\text{bone}}$   & 2.0  & bone-length consistency \\
$\mathcal{L}_{\text{smooth}}$ & 0.01 & latent temporal smoothness \\
$\mathcal{L}_{\text{attn}}$   & 1.0  & attention supervision \\
\bottomrule
\end{tabular}
\end{table}

\subsection{Graph Structural Biases}
\label{app:graph}

Each spatial self-attention layer in the graph-transformer encoder and
decoder adds learned structural biases to the pre-softmax attention
logits.  For a query joint $i$ and key joint $j$, the total bias is:

\begin{equation}
b(i,j)
= \mathbf{q}_i^\top \mathbf{w}^{tq}_{d(i,j)}
+ \mathbf{k}_j^\top \mathbf{w}^{tk}_{d(i,j)}
+ \mathbf{q}_i^\top \mathbf{w}^{eq}_{e(i,j)}
+ \mathbf{k}_j^\top \mathbf{w}^{ek}_{e(i,j)}
+ \mathbf{q}_i^\top \mathbf{w}^{cq}_{c(i,j)}
+ \mathbf{k}_j^\top \mathbf{w}^{ck}_{c(i,j)},
\label{eq:graph_bias}
\end{equation}

where $\mathbf{q}_i, \mathbf{k}_j \in \mathbb{R}^{D/H}$ are per-head
query and key vectors, and $\mathbf{w}^{\cdot}_{\cdot}$ are learned
embedding vectors looked up by three integer indices.

\paragraph{Hop-distance bias ($d(i,j)$).}
$d(i,j)$ is the shortest-path distance between joints $i$ and $j$ on
the kinematic tree, computed via Floyd--Warshall and clamped to
$d_{\max}{=}20$.  Two embedding tables ($\mathbf{w}^{tq}$ and
$\mathbf{w}^{tk}$, each of size $(d_{\max}{+}1)\times D$) provide
query-side and key-side biases that decay with skeleton distance,
encouraging joints that are close in the kinematic tree to attend more
strongly to one another.

\paragraph{Semantic edge-type bias ($e(i,j)$).}
Six semantic categories capture the directed structural relationship
between every pair of joints:

\begin{center}
\small
\begin{tabular}{cl}
\toprule
Type & Condition \\
\midrule
0 & $i{=}j$, non-leaf joint (self) \\
1 & $j$ is the direct parent of $i$ \\
2 & $j$ is a direct child of $i$ \\
3 & $i$ and $j$ share the same direct parent (siblings) \\
4 & all other pairs (no direct relation) \\
5 & $i{=}j$, leaf joint (end-effector) \\
\bottomrule
\end{tabular}
\end{center}

Two embedding tables ($\mathbf{w}^{eq}$, $\mathbf{w}^{ek}$, each
$6\times D$) provide query-side and key-side biases.  Types 1--3
distinguish the immediate kinematic neighbourhood; type 5 specialises
end-effectors, which often carry distinctive motion signatures (hands,
feet, tail tips).

\paragraph{Same-chain binary bias ($c(i,j)$).}
$c(i,j)\in\{0,1\}$ indicates whether joints $i$ and $j$ belong to the
same \emph{kinematic chain}---a maximal linear path in the tree with no
branching nodes between them.  Chain membership is computed once per
skeleton via DFS from the root: single-child nodes continue the current
chain; branching nodes start new chains for each child.  Two embedding
tables ($\mathbf{w}^{cq}$, $\mathbf{w}^{ck}$, each $2\times D$)
encourage joints on the same chain (e.g., all vertebrae of a spine, all
segments of a tail) to form coherent part representations.

All bias terms in \cref{eq:graph_bias} are summed and added to the
standard scaled dot-product scores before softmax, and are shared
across all heads within a block but independent across blocks.
Padded joints (beyond the true skeleton size) are masked to
$-\infty$ before the softmax.

\subsection{Stage 2: MaskGIT Token Generator}

A MaskGIT transformer is trained over the $(T/4){\times}K$ grid of
SAMoR part tokens, conditioned on a text embedding and the target
skeleton.  The main transformer (12 layers, $D{=}512$, 8 heads)
predicts level-0 RVQ code indices; a lightweight residual transformer
(4 layers, same $D$) predicts the remaining $R{-}1$ RVQ levels
conditioned on the decoded level-0 codes.  Text is encoded by a frozen
CLIP text encoder~\cite{radford2021learning} and injected via
cross-attention at every layer.

\paragraph{Training objective.}
Both the main and residual transformers are trained with
cross-entropy loss computed only over masked positions
(unmasked positions are excluded via \texttt{ignore\_index}):
\begin{equation}
\mathcal{L}_{\text{S2}}
= -\!\sum_{m \in \mathcal{M}}
  \log p_\theta\!\bigl(\hat{c}_m \mid c_{\setminus\mathcal{M}},\,\mathbf{y},\,\mathcal{S}\bigr),
\end{equation}
where $\mathcal{M}$ is the set of masked token positions,
$\hat{c}_m$ is the ground-truth code at position $m$,
$\mathbf{y}$ is the CLIP text embedding, and $\mathcal{S}$ is the
skeleton conditioning. Label smoothing $\epsilon{=}0.1$ is applied.

\paragraph{Training details.}
Stage 2 is trained for 300 epochs with batch size 32, learning rate
$10^{-4}$, and a cosine decay schedule with linear warm-up,
on 4$\times$ GH200 96\,GB GPUs ($\sim$12\,h wall-clock).
Tokens are masked with a mixed strategy: each batch randomly applies
either temporal-frame masking (masking all $K$ parts in selected frames)
or random-token masking (independently masking tokens across the
$(T/4){\times}K$ grid), with equal probability.
The masking ratio is sampled from a cosine schedule
$r \sim \cos^{-1}(u) \cdot \frac{2}{\pi}$ for $u\!\sim\!\mathcal{U}[0,1]$,
and a BERT-style corruption is applied: $88\%$ of masked positions
receive a special \textsc{Mask} token, $10\%$ are replaced with a
random code, and $2\%$ are left unchanged.

\paragraph{Inference.}
Tokens are generated iteratively in 18 MaskGIT decoding steps.
At each step, the model predicts all masked positions simultaneously;
the least-confident tokens are re-masked according to a cosine
unmasking schedule, and confident tokens are committed.
Classifier-free guidance~\cite{ho2022cfg} with scale $s{=}4.0$ and
temperature $1.0$ is used at inference; the null condition drops both
the text embedding and the skeleton conditioning.

\subsection{Mesh-to-Animation Frontend}
\label{app:mesh_frontend}

The mesh-to-animation qualitative results in the main paper and
supplementary videos use
Puppeteer~\cite{song2025puppeteer} as the automatic rigging and
skinning frontend.
Given a raw textured mesh, Puppeteer predicts a skeleton and
per-vertex skinning weights; the resulting rigged character is then
passed directly to SAMoR for motion generation or transfer.
Other off-the-shelf auto-rigging systems such as
MagicArticulate~\cite{song2025magicarticulate} and
AnyMate~\cite{deng2025anymate} are compatible with the same pipeline.

% =====================================================================
% C. Canonical Joint-Name Matching
% =====================================================================
\section{Canonical Joint-Name Matching}
\label{app:matching}

Per-joint functional-group labels $y_j \in \{0,\ldots,7\} \cup \{-1\}$ are
derived automatically from raw joint name strings in two stages.

\paragraph{Stage 1: name cleaning.}
Raw names undergo a sequence of normalisation steps: (i)~strip
namespace and vendor prefixes (\texttt{mixamorig:}, \texttt{Bip01\_},
\texttt{BN\_}, etc.) applied iteratively up to three times to handle
nested prefixes; (ii)~convert left/right suffix conventions
(\texttt{\_L}, \texttt{R\_}, \texttt{\_left}, etc.) to a canonical
\texttt{Left}/\texttt{Right} prefix; (iii)~split CamelCase tokens
(\texttt{LeftForeArm} $\to$ \texttt{Left Fore Arm}); (iv)~replace
underscores and dots with spaces and collapse whitespace.

\paragraph{Stage 2: canonical lookup.}
Cleaned names are matched against a dictionary of ${\sim}130$
synonym groups, each mapping surface variants to a canonical name
(e.g., \texttt{upper arm}, \texttt{upperarm}, \texttt{uarm}
$\to$ \texttt{left upper arm}).
Unresolved names undergo progressive fallback: strip residual numeric
suffixes, drop unknown leading tokens, and apply keyword-based regex
patterns (e.g., \texttt{wing}, \texttt{thigh}, \texttt{toe}).
Joints that remain unmatched after all fallback steps receive
$y_j{=}{-1}$ and contribute no terms to the attention supervision loss.

\paragraph{Label coverage.}
Applying the pipeline above, canonical part labels are successfully
assigned to $100\%$ of joints in H3D (standardised MoCap names),
$81\%$ in Zoo (semi-standard creature rigs), and $69\%$ in OXL
(highly diverse artist-authored rigs with vendor-specific naming
conventions). The remaining unlabelled joints are treated as unknown
at both training and inference; name dropout during training ensures
the model does not rely on label availability.

\paragraph{8-slot functional-group mapping.}
Matched canonical names are mapped to one of eight part indices via the
table in \cref{tab:part8}.  Labels are used only during training; at
inference, joints are routed by the learned cross-attention.

\begin{table}[h]
\centering
\caption{Canonical-name-to-part mapping for the $K{=}8$ functional-group slots.
Entries in each cell are canonical names; starred entries cover indexed
variants (\texttt{tail1}--\texttt{tail5}, \texttt{left leg1}--\texttt{left leg4}, etc.).}
\label{tab:part8}
\small
\begin{tabular}{clp{8cm}}
\toprule
\textbf{Index} & \textbf{Part name} & \textbf{Canonical joint names} \\
\midrule
0 & lower torso  & hips, body \\
1 & upper torso  & spine, spine1--3, chest \\
2 & head / neck  & neck, neck1--2, head, jaw, left/right eye, nose,
                   left/right ear, tongue, tongue2,
                   left/right eyebrow \\
3 & left arm     & left clavicle, left shoulder, left upper arm, left elbow,
                   left forearm, left wrist, left hand,
                   left finger0--4, left wing1--4 \\
4 & right arm    & right clavicle, right shoulder, right upper arm, right elbow,
                   right forearm, right wrist, right hand,
                   right finger0--4, right wing1--4 \\
5 & left leg     & left upper leg, left knee, left leg, left horse link,
                   left ankle, left foot, left toe, left toe end,
                   left heel, left leg1--4$^*$ \\
6 & right leg    & right upper leg, right knee, right leg, right horse link,
                   right ankle, right foot, right toe, right toe end,
                   right heel, right leg1--4$^*$ \\
7 & tail         & tail, tail1--5$^*$ \\
\bottomrule
\end{tabular}
\end{table}

% =====================================================================
% D. Metric Definitions
% =====================================================================
\section{Metric Definitions}
\label{app:metrics}

\paragraph{MPJPE (normalised).}
For cross-topology evaluation, all positions are expressed in a
bounding-box normalised coordinate frame (character scaled to unit
bounding box) with the per-frame root joint subtracted.  Mean
Per-Joint Position Error is then:
\begin{equation}
\text{MPJPE} = \frac{1}{T J_\text{real}}
  \sum_{t=1}^{T}\sum_{j\in\mathcal{J}}
  \bigl\|\hat{\mathbf{p}}_j^t - \mathbf{p}_j^t\bigr\|_2,
\end{equation}
where $\mathcal{J}$ indexes non-padded joints and values are reported
in units of $10^{-2}$ (i.e., 1\% of bounding-box diameter).
For HumanML3D, MPJPE is computed on the 22-joint relative-pose
representation in the canonical body frame (root excluded) in
millimetres, following the standard MoMask evaluator.

\paragraph{Bone-length error.}
Squared deviation of predicted bone lengths from ground truth,
averaged over all parent--child pairs and frames:
\begin{equation}
\mathcal{L}_{\text{bone}} = \frac{1}{T |\mathcal{E}|}
  \sum_{t=1}^{T}\sum_{(j,p)\in\mathcal{E}}
  \Bigl(
    \|\hat{\mathbf{p}}_j^t - \hat{\mathbf{p}}_p^t\|_2
    - \|\mathbf{p}_j^t - \mathbf{p}_p^t\|_2
  \Bigr)^2,
\end{equation}
where $\mathcal{E}$ is the set of parent--child edges and $p{=}\mathrm{pa}(j)$.
Reported in the same normalised units as MPJPE.

\paragraph{HumanML3D metrics.}
FID, R-Precision (Top-1/2/3), and MM-Dist are computed using the
standard feature extractor and evaluation protocol from
MoMask~\cite{guo2024momask}, applied to 263-D SMPL motion sequences.
We follow the same evaluation script without modification.

% =====================================================================
% E. Baseline Adaptation Details
% =====================================================================
\section{Baseline Adaptation Details}
\label{app:baselines}

Both fixed-skeleton tokenizers are adapted to variable-$J$ inputs using
a unified padding scheme: skeletons are zero-padded to
$J_{\max}{=}96$ joints, with padded positions masked to $-\infty$ in
attention and excluded from the loss computation.

\paragraph{T2M-GPT (K=1 flat).}
T2M-GPT~\cite{zhang2023generating} is a single-stream VQ-VAE that
encodes each frame as one global token.  We adapt it to arbitrary
skeletons by concatenating all $J$ per-joint position-and-velocity
features into a single flat per-frame feature vector of dimension
$J_{\max}{\times}6$, then applying the original 1D-convolutional
encoder unchanged.  This gives one token per downsampled frame
($K{=}1$ style) with no joint-wise structure.

\paragraph{MoGenTS (K=J per-joint grid).}
MoGenTS~\cite{yuan2024mogents} uses a 2D grid of tokens over time and
joints.  We extend it to arbitrary $J$ by replacing the fixed SMPL-22
joint set with a variable number of joints: each joint receives its own
$d{=}6$ position-and-velocity feature (no rotation), producing a
$T{\times}J$ grid of per-joint tokens ($K{=}J$ style).  The
architecture (1D convolutions along each axis independently) is applied
unchanged; padded joint positions receive zero features and are masked
during attention and loss.

Both baselines are trained on the same unified heterogeneous corpus
as SAMoR with the same data splits, loss functions (position + velocity
smooth-$\ell_1$), and training budget, to ensure a fair comparison.

% =====================================================================
% F. Joint-Name Conditioning Analysis
% =====================================================================
\section{Joint-Name Conditioning Analysis}
\label{app:name_dropout}

We expand on the two name-conditioning rows of the cross-topology
ablation (\emph{labels-only} and \emph{drop name @ inference}) with
the full training-time dropout sweep and a per-source sensitivity
analysis.

\paragraph{Dropout-probability sweep.}
We train SAMoR variants with name-dropout probability
$p\in\{0.0,0.1,0.3,0.5,0.7,0.9,1.0\}$ on the full cross-topology training corpus
and evaluate each checkpoint twice: once with joint-name embeddings
($\textsf{with-name}$), and once with all embeddings replaced by zero
at inference time ($\textsf{drop-name}$).
\cref{tab:name_dropout_sweep} reports both columns, the absolute gap,
and the relative gap. The trade-off forms a clean Pareto frontier:
$p{=}0.0$ achieves the best with-name MPJPE
($2.63\!\times\!10^{-2}$) but is brittle under name removal ($+80\%$
relative gap). SAMoR's default $p{=}0.3$ preserves $95\%$ of the
$p{=}0$ accuracy while reducing the gap from $80\%$ to $16\%$, so a
single checkpoint deploys on both well-named and unnamed rigs.

\begin{table}[h]
  \centering
  \caption{Name-dropout sweep on the SAMoR validation set.
    All values are MPJPE in $10^{-2}$ normalized units (root translation
    excluded, consistent with the HumanML3D convention).
    \emph{Gap (rel.)} is $(\textsf{drop-name}-\textsf{with-name})/\textsf{with-name}$.}
  \label{tab:name_dropout_sweep}
  \begin{tabular}{@{}c c c c c l@{}}
    \toprule
    $p$ & with-name & drop-name & gap (abs.) & gap (rel.) & Notes \\
    \midrule
    $0.0$ & \textbf{2.63} & 4.74 & 2.11 & $80\%$ & best with-name; brittle \\
    $0.1$        & 2.67 & 4.27 & 1.60 & $60\%$ & \\
    $0.3$ & 2.75 & 3.19 & 0.44 & $16\%$ & SAMoR default \\
    $0.5$        & 2.85 & 3.27 & 0.42 & $15\%$ & balanced \\
    $0.7$        & 2.95 & 3.22 & 0.27 & $9\%$  & \\
    $0.9$        & 3.05 & 3.17 & 0.12 & $4\%$  & \\
    $1.0$ & 3.10 & 3.10 & 0 & $0\%$  & labels-only; fully robust \\
    \bottomrule
  \end{tabular}
\end{table}

\paragraph{Per-source sensitivity.}
The drop-name penalty is source-dependent. H3D is most name-sensitive:
at $p{=}0$, removing names inflates H3D MPJPE from
$(3.19\!\to\!7.96)\!\times\!10^{-2}$ ($+150\%$), because the model
exploits the well-vocabularised SMPL joint names. OXL shows an
intermediate $76\%$ jump ($(3.91\!\to\!6.88)\!\times\!10^{-2}$). Zoo is most robust: only
$53\%$ ($(5.30\!\to\!8.13)\!\times\!10^{-2}$), because cross-species diversity forces
reliance on structural priors. At $p{=}1.0$, all three sources become
statistically indistinguishable, indicating zero name dependence.
Bone error and jitter are largely insensitive to dropout, indicating
that name embeddings primarily affect positional accuracy rather than
skeleton consistency or temporal smoothness.

\paragraph{Practical implication.}
The \emph{drop-name @ inference} row uses the $p{=}0.3$ checkpoint
evaluated without names, giving the worst-case behaviour on rigs with
no semantic joint labels. The \emph{labels-only} row trains a separate
$p{=}1.0$ checkpoint and shows the lower bound achievable with zero
name dependence. In deployment we ship the $p{=}0.3$ checkpoint;
unnamed inputs fall through to the drop-name path with the modest cost
reported in the ablation table.

% =====================================================================
% G. Dataset Details
% =====================================================================
\section{Dataset Details}
\label{app:data}

Dataset statistics are summarised in \cref{tab:dataset_stats} of the main paper; here we give additional preprocessing details.

\paragraph{Shared normalization.}
All three sources undergo the same geometric normalization before
training.
Each character is centered and scaled to unit size using
the mesh bounding box.
The up axis is corrected to $+Z$ using foot and leg joint keyword
detection on the skeleton.
The facing direction is aligned to $+Y$ using hip--shoulder or
foot-pair landmarks.
Finally, the skeleton is translated so that the lowest foot joint lies
at $Z{=}0$ (Z-floor alignment).
For OXL and Zoo, the first frame of each motion clip serves as the pose anchor
injected into the encoder and decoder at every graph-transformer layer;
for HumanML3D-compatible experiments, the canonical SMPL rest pose is used instead.

\paragraph{HumanML3D.}
We follow the standard HumanML3D 263-D feature extraction:
root trajectory, relative joint positions, 6D joint rotations, joint
velocities, and binary foot-contact indicators, concatenated per frame.
In addition, we augment the dataset with additional humanoid characters
by fitting SMPL meshes to posed bodies, applying skeleton augmentation
(random perturbations to bone lengths and proportions), and normalizing
the resulting meshes to unit size with the same up-axis and
facing-direction convention as OXL and Zoo.
The official train/val split is used unchanged.

\paragraph{Truebones Zoo.}
Raw FBX files are processed in Blender to extract the skeleton
hierarchy, per-vertex skinning weights, and motion sequences,
which are exported as BVH.
The shared normalization is then applied.
Motions are split at the species level: 54 species for training and
11 species for validation, ensuring zero species overlap between
splits.
Validation species are:
Ant, Buzzard, Chicken, Crocodile, Elephant, Flamingo, Lion, Ostrich,
PolarBearB, Puppy, and SandMouse.
At training time, $T{=}64$ frames are randomly sampled from each
motion; the full motion is used at evaluation.

\paragraph{Objaverse-XL.}
Starting from the raw GLB assets, we run a Blender Python pipeline
that (i)~exports the skinned mesh at the first animation frame as
\texttt{object.obj} (with UV and texture maps), and
(ii)~exports the full animation as a BVH in which all joint rotations
are expressed relative to the first frame
($\mathbf{R}_t^{\mathrm{rel}} = \mathbf{R}_0^{-1}\mathbf{R}_t$).
The shared normalization is applied after export.

The resulting assets pass through a multi-stage curation pipeline:
\begin{enumerate}[leftmargin=*,topsep=2pt,itemsep=1pt]
  \item \textbf{Rigging validity.}
    Characters whose skeleton root bone lies outside the mesh bounding
    box are discarded, along with a manually curated exclusion list of
    visually confirmed broken rigs.
  \item \textbf{Inside-mesh filter.}
    We voxelize the mesh at $64^3$ resolution and discard any character
    for which more than 50\% of skeleton joints fall outside the mesh
    volume, indicating a misaligned or broken rig.
  \item \textbf{Near-static filter.}
    Characters whose maximum per-joint rotation standard deviation
    across time falls below a threshold, combined with low joint
    displacement, are removed as effectively pose-only assets.
  \item \textbf{Repetition filter.}
    Pairs of characters sharing identical joint counts, identical parent
    connections, and near-identical joint positions on five randomly
    sampled frames are treated as duplicates; all but one are discarded.
  \item \textbf{Joint count.}
    Characters with fewer than 5 or more than 96 joints are excluded.
  \item \textbf{Short clip filter.}
    Motion clips shorter than 16 frames at 20\,fps are discarded.
  \item \textbf{Out-of-bounds filter.}
    Before training, clips in which any joint position exceeds
    $3.0$ normalised units from the origin after shared normalisation
    are removed, as these indicate residual normalisation failures.
\end{enumerate}

After curation, the remaining characters are split at the
character level (zero character overlap between train and val).

\paragraph{OXL character type distribution.}
We generate one motion caption per character by rendering a
$4{\times}4$ keyframe strip (front/side views $\times$ mesh/skeleton)
and prompting Gemini Flash~\cite{team2024gemini} to classify the
character type and produce short, medium, and detailed captions.
Characters for which Gemini returns no type label (API non-responses)
are assigned the \textsc{Generic} type.
\Cref{tab:oxl_types} reports the resulting type distribution.

\begin{table}[h]
\centering
\caption{OXL character type distribution over all 71,174 curated
  characters. Characters without a Gemini-assigned label are
  counted as \textsc{Generic}.}
\label{tab:oxl_types}
\small
\begin{tabular}{@{}lrr@{}}
\toprule
Type & Characters & \% \\
\midrule
\textsc{Humanoid}    & 24,596 & 34.6 \\
\textsc{Generic}     & 31,374 & 44.1 \\
\textsc{Biped Other} &  5,866 &  8.2 \\
\textsc{Quadruped}   &  4,592 &  6.5 \\
\textsc{Winged}      &  3,099 &  4.4 \\
\textsc{Multi-limb}  &  1,647 &  2.3 \\
\midrule
\textbf{Total}       & 71,174 & 100.0 \\
\bottomrule
\end{tabular}
\end{table}

% =====================================================================
% H. Pose Anchor Design
% =====================================================================
\section{Pose Anchor Design}
\label{app:anchor}

SAMoR injects a per-character \emph{pose anchor} into every
graph-transformer layer to condition the encoder and decoder on
skeleton geometry.
When a canonical rest pose is available---as in HumanML3D, where all
characters share the SMPL rest pose---we use it directly.
For heterogeneous cross-topology assets (OXL, Zoo), rest poses are
not consistently defined across rigs: different exporters and artists
produce T-poses, A-poses, or arbitrary bind poses that are not
comparable across species or asset styles.
Using an ill-defined rest pose as the anchor would introduce
inconsistent geometry conditioning, reducing the model's ability to
generalize across the diverse topologies in our corpus.

We therefore use the \emph{first frame} of each motion clip as the
pose anchor for OXL and Zoo.
This is a principled choice: the first frame is always available,
is free of exporter inconsistencies, and---because different characters
start from different first-frame configurations---encourages the model
to learn motion representations that are robust to diverse starting
poses rather than overfitting to a fixed rest-pose convention.
For reconstruction, the first frame belongs to the input clip and is
available at test time; since MPJPE is computed in per-frame
root-relative coordinates, the absolute starting pose does not
inflate the reconstruction metric.
At transfer or generation time the pose anchor is drawn exclusively
from the \emph{target} rig and carries no information from the source
motion, so cross-topology evaluation is free of motion-cue leakage
through the anchor.

% =====================================================================
% A. Additional Qualitative Results
% =====================================================================
\section{Additional Qualitative Results}
\label{app:qual}

We now provide additional qualitative evaluations regarding the part assignment of SAMoR, followed by end-to-end mesh animation and part composition.

\begin{figure*}[!htbp]
  \centerline{\includegraphics[width=\linewidth]{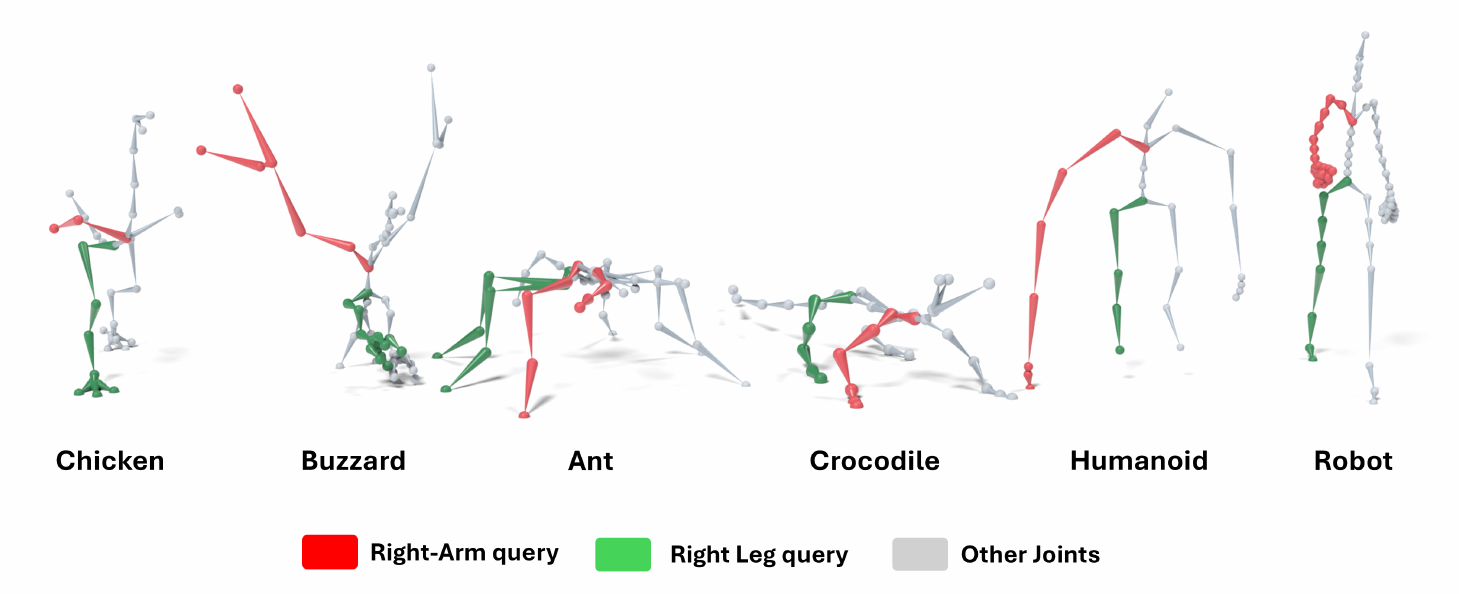}}
  \caption{\textbf{Functional-query alignment across skeleton topologies.}
    Dominant joint assignment of SAMoR's right-arm-like (red) and
    right-leg-like (green) queries on six held-out characters
    (Chicken, Buzzard, Fire Ant, Crocodile, Humanoid, Robot)
    at inference time (no joint names provided). Each query localizes
    to functionally corresponding appendage chains---forelimbs, wings,
    and leading insect legs for the arm-like query; hindlimbs and
    walking-leg pairs for the leg-like query---showing that
    $\mathcal{L}_\text{attn}$ (\cref{sec:method:attn}) induces
    functional group routing that generalizes across biological,
    non-biological, and mechanical rigs.}
  \label{fig:partattn}
\end{figure*}

\paragraph{Part assignment visualizations.}
\cref{fig:partattn} shows inference-time part assignment of SAMoR's
right-arm and right-leg queries on six held-out characters with no joint
names provided. The model routes functionally corresponding appendages
to corresponding functional slots across biological (Chicken, Buzzard, Crocodile),
non-biological (Robot), insect (Fire Ant), and humanoid rigs.

\begin{figure*}[p]
  \centering
  \includegraphics[width=\linewidth,height=0.86\textheight,keepaspectratio]{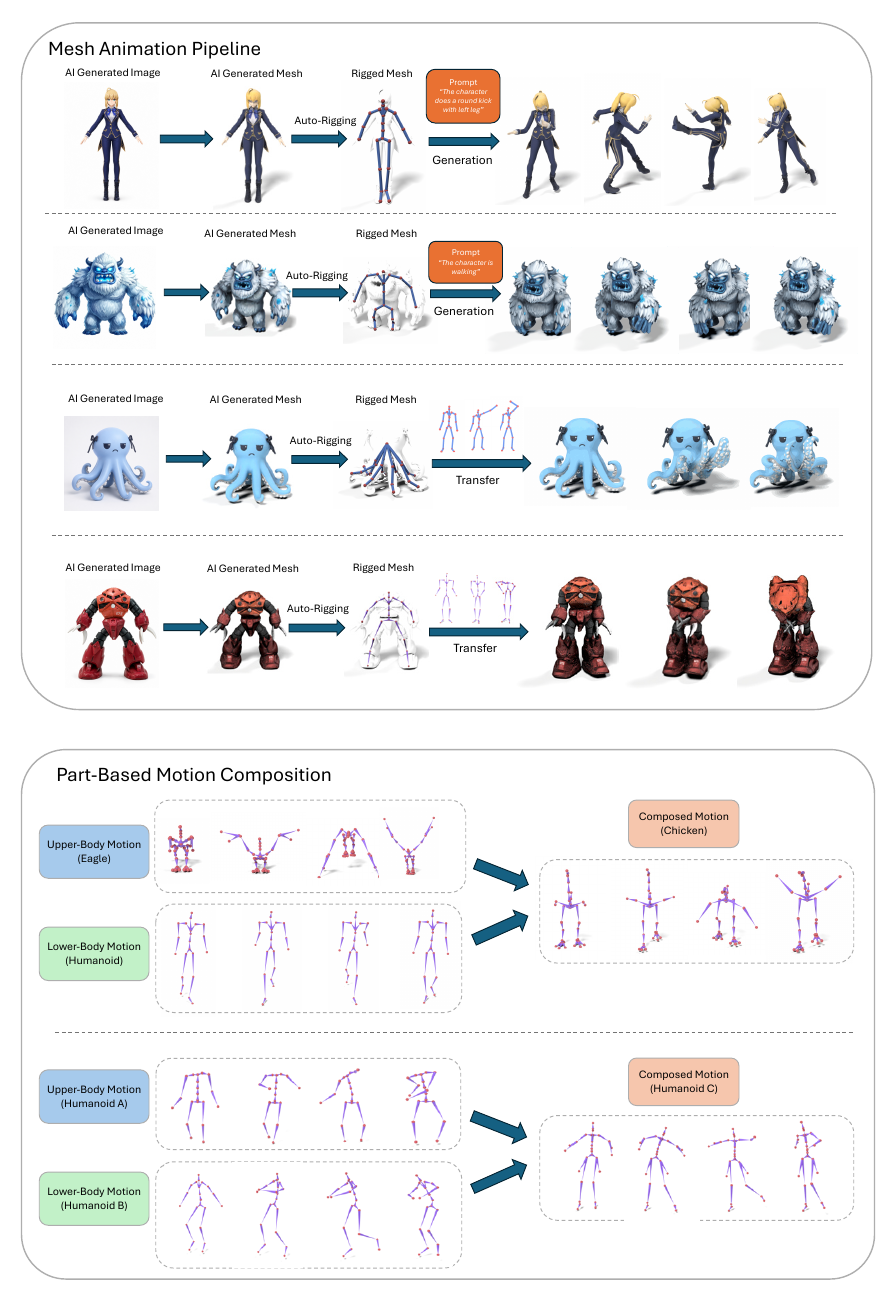}
  \caption{\textbf{End-to-end mesh animation and part-based motion composition.}
    \emph{Top --- Mesh Animation Pipeline:} from a single AI-generated image we
    reconstruct a 3D mesh, auto-rig it, and drive it with SAMoR via
    text-conditioned generation and cross-topology motion transfer.
    \emph{Bottom --- Part-Based Motion Composition:} SAMoR's per-part tokens
    let us compose a new motion by pairing the upper-body part tokens of one
    source motion with the lower-body part tokens of another, yielding a
    coherent blended motion.
    Video versions of all examples are on our \href{\projecturl}{project website}.}
  \label{fig:supp_qual}
\end{figure*}

\ifdefined\arxivmode
% arXiv-only: extended qualitative figures live in THIS section; videos on the project website.

\paragraph{End-to-end mesh animation and part composition.}
\cref{fig:supp_qual} shows two complete mesh-animation pipelines---driving an
AI-generated, auto-rigged mesh by text-conditioned generation and by
cross-topology motion transfer---together with a part-based motion composition
example that blends the upper- and lower-body part tokens of different source
motions. Video versions of all examples are on our
\href{\projecturl}{project website}.
\else
\paragraph{Extended results (video).}
Cross-topology motion transfer, text-conditioned generation, part-wise
token editing, and mesh LBS animation results are provided as an HTML
supplementary page; open \texttt{index.html} in a browser to view all videos.
\fi

\end{document}